\documentclass{article}
\usepackage{times}
\usepackage{subcaption}
\usepackage{url}
\usepackage{hyperref}
\usepackage{graphicx}
\graphicspath{ {./images/} }
\usepackage{color,soul}

\usepackage{wrapfig}
\usepackage{subcaption}
\usepackage{cleveref}
\usepackage{url}
\usepackage{amsmath}
\usepackage{float}

\usepackage [english]{babel}
\usepackage [autostyle, english = american]{csquotes}
\MakeOuterQuote{"}

\usepackage{amsmath,amsfonts,bm}

\def\eqref#1{equation~\ref{#1}}
\def\1{\bm{1}}

\def\mW{{\bm{W}}}

\DeclareMathAlphabet{\mathsfit}{\encodingdefault}{\sfdefault}{m}{sl}
\SetMathAlphabet{\mathsfit}{bold}{\encodingdefault}{\sfdefault}{bx}{n}

\newcommand{\R}{\mathbb{R}}

\newcommand{\comment}[1]{}

\usepackage[accepted]{icml2019}

\icmltitlerunning{Dynamic Planning Networks}

\begin{document}

\twocolumn[
\icmltitle{Dynamic Planning Networks}


\icmlsetsymbol{equal}{*}

\begin{icmlauthorlist}
\icmlauthor{Norman Tasfi}{uwo}
\icmlauthor{Miriam Capretz}{uwo}
\end{icmlauthorlist}

\icmlaffiliation{uwo}{Department of Electrical and Computer Engineering, University Of Western Ontario, London, Ontario, Canada}

\icmlcorrespondingauthor{Norman Tasfi}{ntasfi@uwo.ca}

\icmlkeywords{Machine Learning, ICML}

\vskip 0.3in
]



\printAffiliationsAndNotice % otherwise use the standard text.

\begin{abstract}
We introduce Dynamic Planning Networks (DPN), a novel architecture for deep reinforcement learning, that combines model-based and model-free aspects for online planning.
Our architecture learns to dynamically construct plans using a learned state-transition model by selecting and traversing between simulated states and actions to maximize information before acting.
In contrast to model-free methods, model-based planning lets the agent efficiently test action hypotheses without performing costly trial-and-error in the environment.
DPN learns to efficiently form plans by expanding a single action-conditional state transition at a time instead of exhaustively evaluating each action, reducing the required number of state-transitions during planning by up to 96\%.
We observe various emergent planning patterns used to solve environments, including classical search methods such as breadth-first and depth-first search.
DPN shows improved data efficiency, performance, and generalization to new and unseen domains in comparison to several baselines.
\end{abstract}

\section{Introduction}

The central focus of reinforcement learning (RL) is the selection of optimal actions to maximize the expected reward in an environment where the agent must rapidly adapt to new and varied scenarios.
Various avenues of research have spent considerable efforts improving core axes of RL algorithms such as performance, stability, and sample efficiency. 
Significant progress on all fronts has been achieved by developing agents using deep neural networks with model-free RL \citep{mnih2015human, mnih2016asynchronous, schulman2015trust, schulman2017proximal, openai_2018}; showing model-free methods efficiently scale to high-dimensional state space and complex domains with increased compute. 
Unfortunately, model-free policies are often unable to generalize to variances within an environment as the agent learns a policy which directly maps environment states to actions. 

\begin{figure*}[t!]
    \begin{subfigure}[h]{0.50\textwidth}
    \includegraphics[width=\linewidth,scale=0.25]{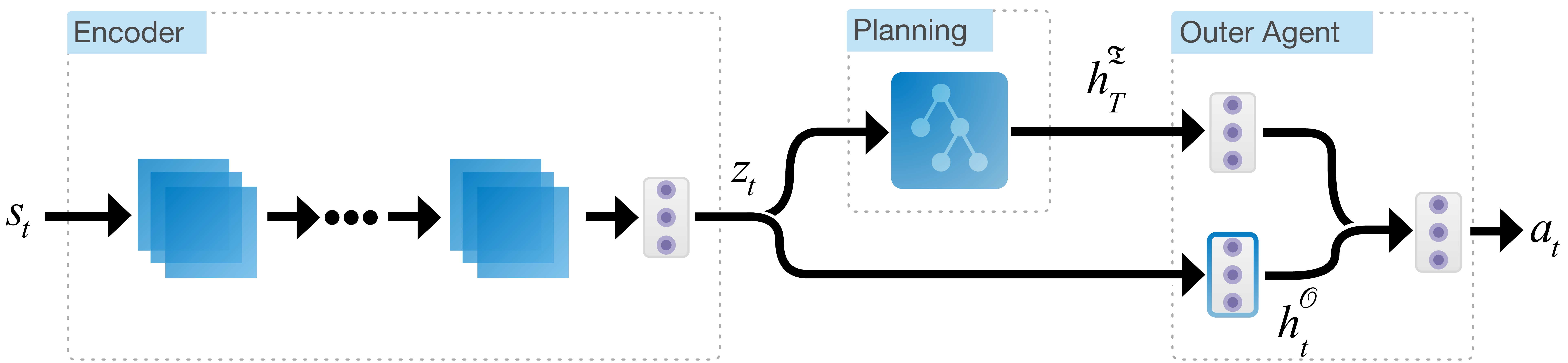}
    \caption{Architecture Overview.}
    \label{fig:arch_overview}
    \end{subfigure}
    \hspace*{\fill}
    \begin{subfigure}[h]{0.50\textwidth}
    \includegraphics[width=\linewidth,scale=0.25]{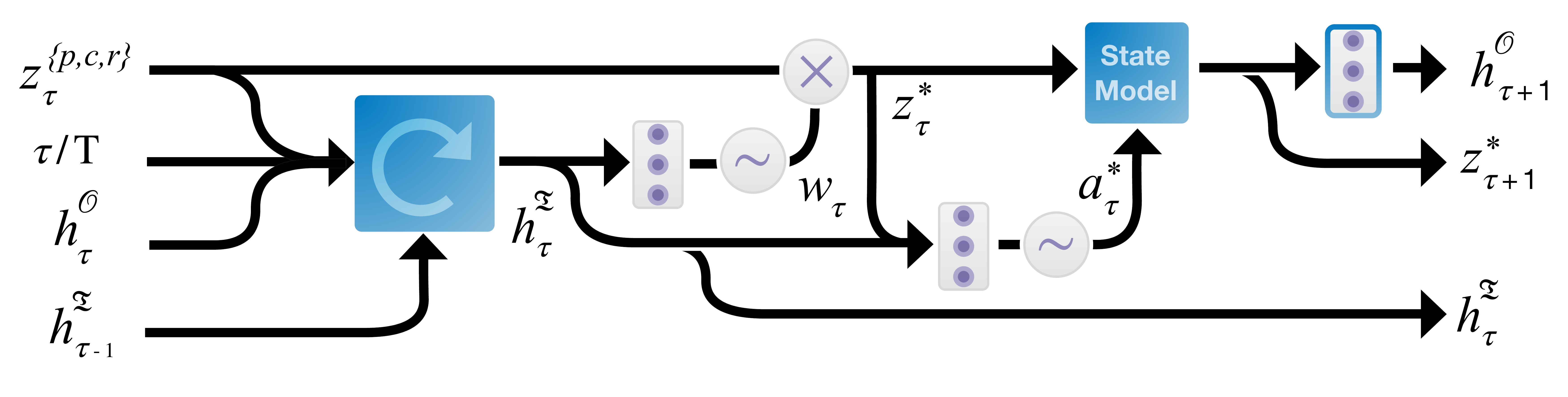}
    \caption{A Planning Step.}
    \label{fig:arch_planning_step}
    \end{subfigure}
    \caption{Dynamic Planning Network Architecture. Encoder is comprised of several convolutional layers and a fully-connected layer. Planning occurs for $\tau=1,...,T$ steps using the IA and state-transition model. The result of planning is sent to the outer agent before an action $a_t$ is chosen. The fully-connected layer within the outer agent, outlined in blue, is used by the planning process.
    b) A single planning step $\tau$. Inner Agent, shown as the blue box with a recursive arrow, performs a step of planning using the state-transition model. Circles containing $\times$ indicate multiplication and circles with $\sim$ indicate sampling from the Gumbel Softmax distribution.}
\end{figure*}

A favorable approach to improving generalization is to combine an agent with a learned environment model, enabling it to reason about its environment. This approach, referred to as model-based RL learns a model from past experience, where the model usually captures state-transitions, $p(s_{t+1}| s_t,a_t)$, and might also learn reward predictions $p(r_{t+1}|s_t,a_t)$. 
Usage of learned state-transition models is especially valuable for planning, where the model predicts the outcome of proposed actions, avoiding expensive trial-and-error in the actual environment -- improving performance and generalization. 
This contrasts with model-free methods which are explicitly trial-and-error learners \citep{suttonRL2nd}.
Historically, applications have primarily focused on domains where a state-transition model can be easily learned, such as low dimensional observation spaces \citep{peng1993efficient, deisenroth2011pilco, levine2014learning}, or where a perfect model was provided \citep{coulom2006efficient, silver2016mastering} -- limiting usage. 
Furthermore, application to environments with complex dynamics and high dimensional observation spaces has proven difficult as state-transition models must learn from agent experience, suffer from compounding function approximation errors, and require significant amounts of samples and compute \citep{OhGLLS15, chiappa2017recurrent, guzdial2017game}. Fortunately, recent work has overcome the aforementioned difficulties by learning to interpret imperfect model predictions \citep{weber2017imagination} and learning in a lower dimensional state space \citep{farquhar2017treeqn}.

Planning in RL has used state-transition models to perform simulated trials with various styles of state traversal such as: recursively expanding all available actions per state for a fixed depth \citep{farquhar2017treeqn}, expanding all actions of the initial state and simulating forward for a fixed number of steps with a secondary policy \citep{weber2017imagination}, or performing many simulated rollouts with each stopping when a terminal state is encountered \citep{silver2016mastering}. 
An issue arises within simulated trials when correcting errors in action selection, as actions can either be undone by wasting a simulation step, using the opposing action, or are irreversible, causing the remaining rollout steps to be sub-optimal in value. 
Ideally, the agent can step the rollout backwards in time thereby undoing the poor action and choosing a better one in its place. 
Additionally, during rollouts the agent is forced to either perform a fixed number of steps or continue until a terminal state has been reached; when ideally a rollout can terminate early if the agent decides the path forward is of low value. 

In this paper, we propose an architecture that learns to dynamically control a state-transition model of the environment for planning. 
By doing so, our model has greater flexibility during planning allowing it to efficiently adjust previously simulated actions. 
We demonstrate improved performance against both model-free and planning baselines on varied environments.

The paper is organized as follows: Section \ref{sec:dynamic_planning_network} covers our architecture and training procedure, Section \ref{sec:related_worked} covers related work, Section \ref{sec:experiments} details the experimental design used to evaluate our architecture, and in Section \ref{sec:results} we analyze the experimental results of our architecture.

\section{Dynamic Planning Network}
\label{sec:dynamic_planning_network}

In this section, we describe DPN, a novel planning architecture for deep reinforcement learning (DRL). 
We first discuss the architecture overview followed by the training procedure. Steps taken in the environment use subscript $t$ and steps taken during planning use subscript $\tau$.
%We provide additional motivation behind the architecture in Appendix \ref{app:motivating}.

\subsection{DPN Architecture}

The architecture is comprised of an inner agent, an outer agent, a shared encoder, and a learned state-transition model. Figure \ref{fig:arch_overview} illustrates a high-level diagram of the DPN architecture.
The outer agent (OA) is a feed-forward network and the inner agent (IA) is based on a recurrent neural network (RNN).
The architecture interacts with the environment by observing raw environment states $s_t \in \mathcal{S}$ and outputting actions $a_t \in \mathcal{A}$ via OA. 
However, before OA outputs an action $a_t$ the IA performs $\tau=1,...,T$ steps of planning by interacting with an internal simulated environment; where this simulated environment is defined by the state-transition model and sub-section of OA's network.
Selection of an action $a_t$ by OA uses the final hidden state $h^{\mathcal{I}}_T \in \R^{1 \times h^i}$ of IA and an embedding of the current state $z_t \in \R^{1 \times z}$. 
The objective of IA is to maximize the total planning utility provided to the OA; where planning utility, given in Equation \ref{eqn:utility}, measures the value of the new state $z_{\tau+1}$ and the change in the OA's hidden state if it were to have undergone a state transition from $z_{\tau}$ to $z_{\tau+1}$.

\begin{equation}
	\label{eqn:utility}
	\begin{gathered}
	\mathcal{U}_{\tau}(h^{\mathcal{O}}_{\tau+1}, h^{\mathcal{O}}_{\tau}, z_{\tau}) = V(z_{\tau+1}) + \mathbf{D}[h^{\mathcal{O}}_{\tau+1}, h^{\mathcal{O}}_{\tau}]
	\end{gathered}
\end{equation}

where $z_{\tau+1}$ is the state transitioned to after performing an action $a_{\tau}$ in state $z_{\tau}$, $h^{\mathcal{O}}_{\tau} \in \R^{1 \times h^o}$ and $h^{\mathcal{O}}_{\tau+1}$ are the hidden states of OA after perceiving the current state $z_{\tau}$ and state transitioned to $z_{\tau+1}$ respectively, $\mathbf{D}$ is a distance measure, and $V(z_{\tau+1})$ is the value OA assigns the next state $z_{\tau+1}$. 
After empirical evaluation of different distance functions, such as L2, Cosine distance, and KL, we found the L1 distance function to be the most performant.

The planning utility formulation is similar to previous work on intrinsic motivation \citep{
plappert2017parameter, chentanez2005intrinsically}. 
Where agent rewards are usually defined as $r^e_t + r^i_t$ with $r^e_t$ representing the external reward given by the environment and $r^i_t$ is an internally generated reward. 
In the case of Equation \ref{eqn:utility}, external reward is provided by the OA's value function and the internal reward is defined by the L1 distance between OA's internal representation of $z_{\tau}$ and $z_{\tau+1}$.

Therefore, to maximize planning utility for OA during each planning step $\tau$, IA must select appropriate simulated-states $z^{*}_{\tau}$ and actions $a^{*}_{\tau}$. 
A simulated-state $z^{*}_{\tau}$ is selected from one of three embedded states tracked during planning: the previous $z^p_{\tau}$, current $z^c_{\tau}$, and root states $z^r_{\tau}$; with the triplet written as $z^{\{p,c,r\}}_{\tau}$ for convenience.
Initially, $z^{\{p,c,r\}}_{\tau=0}$ is set to an embedding $z_t$ produced by the encoder of the initial raw state $s_t$ as $z_{\tau=0} = encoder(s_t)$. The encoder is comprised of a series of convolutional layers specific to each environment.
Before planning begins, OA's hidden state is updated using $z_{\tau=0}$:

\begin{equation}
\label{eqn:outer_h_update}
h^{\mathcal{O}}_{\tau} = W^{zh}z_{\tau}
\end{equation}

where $\mW^{zh} \in \R^{z \times h^o}$ is a learnable parameter of OA with biases omitted.
Within this work, we consider the intermediate activation from the OA, a feed-forward network, as a hidden state.
The simulated-action $a^{*}_{\tau}$ mirrors those available to OA in the environment, such that $a^{*}_{\tau} \in \mathcal{A}$.

Empirically, we found that using the same policy for planning and acting caused poor performance. 
We hypothesize that the optimal policy for planning is inherently different from the one required for optimal control in the environment; as during planning, a bias toward exploration might be optimal.

\subsection{A Planning Step}
\label{ssec:a_planning_step}

At each planning step $\tau$, shown in Figure \ref{fig:arch_planning_step}, the IA selects a simulated-state $z^{*}_{\tau}$ and action $a^{*}_{\tau}$ by considering the previous hidden state $h^{\mathcal{I}}_{\tau-1}$, the triplet of embedded states $z^{\{p,c,r\}}_{\tau}$, a scalar representing the current planning step $\tau/T$, and OA's hidden state $h^{\mathcal{O}}_{\tau}$ given $z^c_{\tau}$.
The information is concatenated together forming a context and is fed into IA, a recurrent network producing an updated hidden state $h^{\mathcal{I}}_{\tau} \in \R^{1xh^i}$.
The updated hidden state is used to select the simulated-state $z^{*}_{\tau}$ by multiplying $z^{\{p,c,r\}}_{\tau}$ with a 1-hot encoded weight $w_{\tau} \in \{0, 1\}^{1 \times 3}$ sampled from the Gumbel-Softmax distribution, \textit{G}:

\begin{equation}
\label{eqn:inner_zselection}
\begin{gathered}
	w_{\tau} \sim \textit{G}(\mW^{h3}h^{\mathcal{I}}_{\tau})\\
	z^{*}_{\tau} = w_{\tau}[z^{p}_{\tau}, z^{c}_{\tau}, z^{r}_{\tau}]
\end{gathered}
\end{equation}

where $\mW^{h3} \in \R^{h^i \times 3}$ is a learnable parameter belonging to IA and \textit{G} is the Gumbel-Softmax distribution (GSD). Where the GSD is a continuous relaxation of the discrete categorical distribution giving access to differentiable discrete variables \citep{jang2016categorical, maddison2016concrete}.
Empirically, we found that using a 1-hot encoding for the weight $w_{\tau}$ gives greater performance than a softmax activation. 
Therefore, we used GSD in place of softmax activations throughout our architecture. 
Next, the simulated-action $a^{*}_{\tau} \in \{0,1\}^{1 \times \mathcal{A}}$, is sampled as follows:

\begin{equation}
	\label{eqn:inner_aselection}
	a^{*}_{\tau} \sim \textit{G}(\mW^{azh}[z^{*}_{\tau}, h^{\mathcal{I}}_{\tau}])
\end{equation}

where $\mW^{azh} \in \R^{a \times z+h^i}$ is a learnable parameter of IA. In Equation \ref{eqn:inner_aselection} the selected simulated-state $z^{*}_{\tau}$ and IA's hidden state $h^{\mathcal{I}}_{\tau}$ are concatenated, passed through a linear layer, and used as logits for GSD. 
Then, with the selected simulated-state $z^{*}_{\tau}$ and simulated-action $a^{*}_{\tau}$, we produce the next state $z_{\tau+1}$ using the state-transition model, defined as:

\begin{equation}
\label{eqn:st_zprime}
\begin{gathered}
	z^{\prime} = z_{\tau} + tanh(\mW^{zz} z^{*}_{\tau}) \\
	z^{\prime\prime} = z^{\prime} + tanh((a^{*}_{\tau} \mW^{azz}) z^{\prime})\\
    z^{*}_{\tau+1} = z_{\tau} + z^{\prime\prime}
\end{gathered}
\end{equation}

where $\mW^{zz} \in \R^{z \times z}$ and $\mW^{azz} \in \R^{\mathcal{A} \times z \times z}$ are learnable parameters of the state-transition model. 
We parameterize each available action in $\mathcal{A}$ with a learned weight matrix that carries information about the effect of taking an action $a^{*}_{\tau} \in \R^{1 \times \mathcal{A}}$. 
We use the same state-transition model presented by \citet{farquhar2017treeqn}.
Finally, the three embedded states are updated as: $z^{p}_{\tau+1} = z^{c}_{\tau}$, $z^{r}_{\tau+1} = z_{\tau=0}$, and $z^c_{\tau+1} = z^{*}_{\tau+1}$. 

\subsection{Action selection}
\label{ssec:action_selection}

The IA repeats the process defined in Section \ref{ssec:a_planning_step} of selecting $z^{*}_{\tau}$ and $a^{*}_{\tau}$ for $T$ steps before finally emitting a final hidden state $h^{\mathcal{I}}_{T}$ summarizing the result of planning.
The OA uses IA's final hidden state $h^{\mathcal{I}}_{T}$ and its initial hidden state $h^{\mathcal{O}}_{\tau=0}$ to select an action $a_t$:

\begin{equation}
	a_t = \mW^{ah}tanh(\mW^{hh}h^{\mathcal{I}}_{T} + h^{\mathcal{O}}_{\tau=0})
\end{equation}

where $\mW^{hh} \in \R^{h^{o} \times h^{i}}$ and $\mW^{ah} \in \R^{\mathcal{A} \times h^{o}}$ are learnable parameters of OA.
Finally, the hidden state of the IA is reset.

\subsection{Tree Interpretation}

\begin{figure}[h!]
    \includegraphics[width=0.5\textwidth]{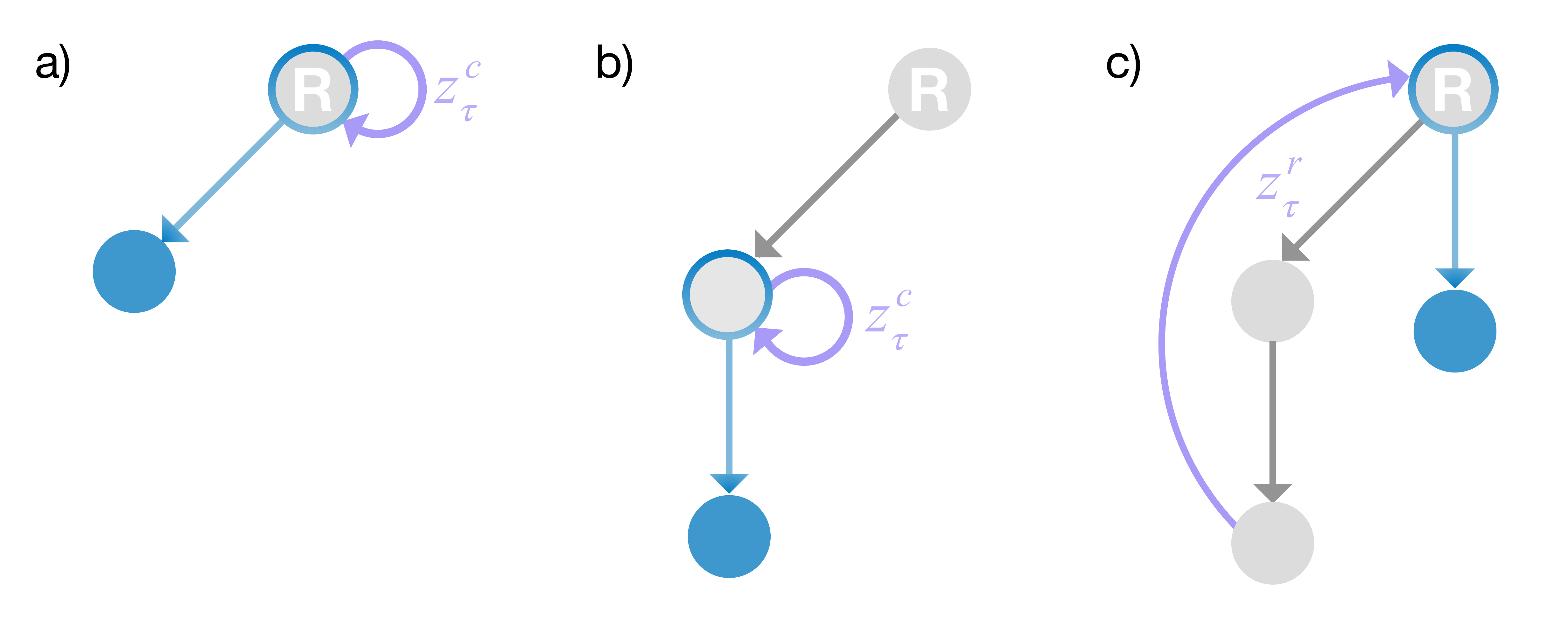}
    \caption{Example of dynamic tree construction during planning.}
    \label{fig:tree_planning}
\end{figure}

The planning process can be interpreted as dynamically expanding a state-action tree, illustrated in Figure \ref{fig:tree_planning}, where all edges and vertexes are chosen by IA to maximize the total planning utility provided to OA.

With simulated-state selection $z^{*}_{\tau}$, using $w_{\tau}$, IA controls which node in the tree is expanded further: the parent node ($z^p$), the root node ($z^r$), or the current node ($z^c$). 
While action selection $a^{*}$ chooses the branching direction, exploring the embedded state space using the state-transition model.

The illustration of a constructed tree in a fictional environment is shown in Figure \ref{fig:tree_planning}. 
State selections are shown in light purple, and state transitions with an action, using the state-transition model, are shown as blue. 
The source state is shown as a grey circle with a blue outline and the transitioned state is shown as a fully blue circle. 
In this example, there are three actions, each corresponding to their graphical representation: left, right, and down. 
The root state is marked with an "R". 
An example of a possible tree construction for T=3 steps of planning: a) step $\tau$ = 1, IA selects the current state $z^c$ and transitions to a new state with action "left"; b) step $\tau$ = 2, the IA selects the current state $z^c$ and "down" action; and step $\tau=3$ c) IA selects the root state $z^c$ and "down" action.

\subsection{Training Procedure}
\label{sec:training_procedure}

We trained the parameters of DPN using A2C, a single threaded version of asynchronous
advantage actor-critic (A3C) \citep{mnih2016asynchronous}. 
The OA was trained as an actor-critic, with policy and value networks, while the IA was trained as a policy network only. As defined in Equation \ref{eqn:utility}, IA used OA's value network as part of its target.
A Huber loss was added to the loss function to perform state-grounding of the state-transition model between the current state $z_{t}$, action $a_t$, and $z_{t+1}$ which we denoted with $\mathcal{L}_{\mathcal{Z}}$. Combining our losses, the architecture is trained using the following gradient:
\begin{equation}
    \Delta\theta = \nabla_{\theta} \mathcal{L}_{\mathcal{O}} +
    \nabla_{\theta} \mathcal{L}_{\mathcal{I}} +
    \lambda\nabla_{\theta}\mathcal{L}_{\mathcal{Z}}
    - \beta\nabla_{\theta}H
\end{equation}

where $\mathcal{L}_{\mathcal{O}}$ is the OA loss (policy and value), $\mathcal{L}_{\mathcal{I}}$ is the IA policy loss, $\lambda$ controls the state-grounding loss, $H$ is the entropy regularizer computed for OA and IA, and $\beta$ is hyperparameter tuning entropy maximization of all policies; we used the same $\beta$ value for each policy. The losses $\mathcal{L}_{\mathcal{O}}$ and $\mathcal{L}_{\mathcal{Z}}$ are computed over all parameters; while $\mathcal{L}_{\mathcal{I}}$ and its entropy regularizer losses are computed with respect to only IA's parameters.
We perform updates to IA in this way as to stop IA from cheating by modifying the parameters of the OA that define its reward via $\mathbf{D}[h^{\mathcal{O}}_{\tau+1}, h^{\mathcal{O}}_{\tau}]$ and $V(z_{\tau+1})$ within the planning utility.

We used 16 workers with RMSprop optimizer \cite{Tieleman2012} across all environments. 

\begin{figure*}[t]
    \begin{subfigure}[b]{0.50\textwidth}
        \begin{subfigure}{\textwidth}
            \includegraphics[width=\textwidth,trim={9.5cm 0 0 0},clip]{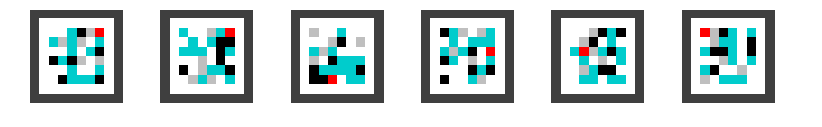}
            \caption{Push Environment Samples.}
            \label{fig:push_environment_samples}
        \end{subfigure}\\
        \vspace*{\fill}
        \begin{subfigure}{\textwidth}
            \vspace{0.2cm}
            \centering
            \begin{tabular}{c|c}
            \hline
            Model & Avg. Reward \\
            \hline
            A2C  &  5.62 \\
            ATreeC-1 &  6.68 \\
            \textit{DPN-T3} &  \textbf{6.99} \\
            DQN     & 3.96 \\
            TreeQN-3     & 5.08 \\
            \end{tabular}
            \caption{Model Performance.}
            \label{table:model_perf}
        \end{subfigure}
    \end{subfigure}
    \begin{subfigure}[b]{0.5\linewidth}
        \centering
        \raisebox{0.05\height}{
            \includegraphics[width=\textwidth]{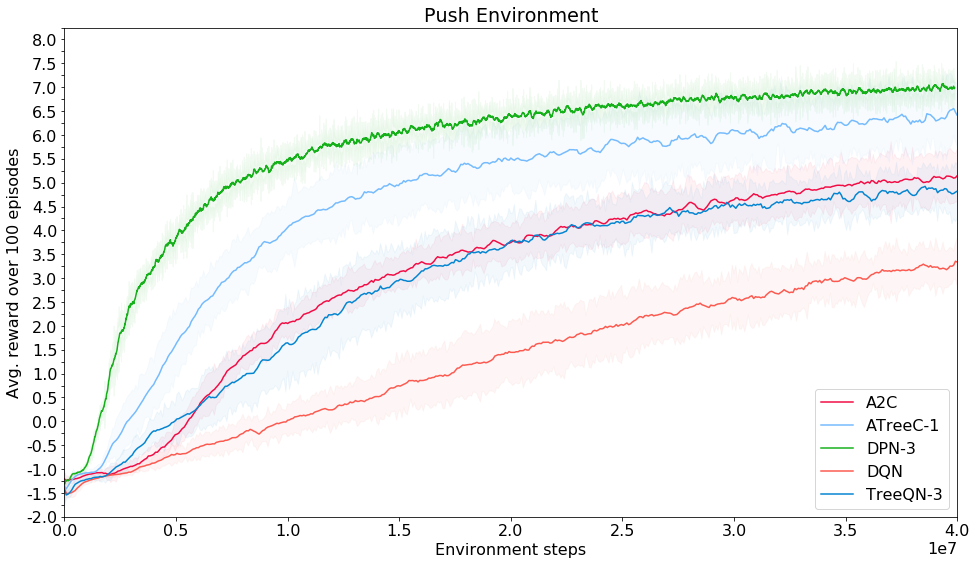}
        }
        \caption{Training Curve.}
        \label{fig:push_training_curve}
    \end{subfigure}
    \caption{
    \textit{Push Environment.}
    \textit{a)} Randomly generated samples of the Push environment. Each square's coloring represents a different entity: the agent is shown as red, boxes as aqua, obstacles as black, and goals as grey. The outside of the environment, not visible to the agent, is shown as a black border around the map. \textit{b)} The performance of each model where \textit{Avg. Reward} is the average of the last 1000 episodes of training. \textit{c)} Training curves with DPN compared to various baselines on  Push environment.
    }
    \label{fig:push_env}
\end{figure*}

\section{Related Work}
\label{sec:related_worked}
Various efforts have been made to combine model-free and model-based methods, such as the \textit{Dyna-Q} algorithm \citep{sutton1991dyna} that learns a model of the environment and uses this model to train a model-free policy.
Originally applied in the discrete setting, \citet{gu2016continuous} extended Dyna-Q to continuous control. 
In a similar spirit to the Dyna algorithm, recent work by \citet{ha2018world} combined data generated from a pre-trained unsupervised model with evolutionary strategies to train a policy. 
However, none of the aforementioned algorithms use the learned model to improve the online performance of the policy and instead use the model for offline training.
Therefore, the learned models are typically trained with a tangential objective to that of the policy such as a high-dimensional reconstruction. 
In contrast, our work learns a model in an end-to-end manner, such that the model is optimized for its actual use in planning.

\citet{guez2018learning} proposed MCTSnets, an approach for learning to search where they replicate the process used by MCTS. MCTSnets replaces the traditional MCTS components by neural network analogs. The modified procedure evaluates, expands, and back-ups a vector embedding instead of a scalar value. The entire architecture is end-to-end differentiable.

\citet{tamar2016value} trained a model-free agent with an explicit differentiable planning structure, implemented with convolutions, to perform approximate on-the-fly value iteration.
As their planning structure relies on convolutions, the range of applicable environments is restricted to those where state-transitions can be expressed spatially.

\citet{PascanuLVHBRRWW17} implemented a model-based architecture comprised of several individually trained components that learn to construct and execute plans.
Their work used a single policy for planning and acting while DPN has a separate policy for each mode.
The policy used in DPN for planning also selects which state to plan from while \citet{PascanuLVHBRRWW17} used a separate specialized policy.
They examine performance on Gridworld tasks with single and multi-goal variants but on an extremely limited set of small maps.

\citet{vezhnevets2016strategic} proposed a method which learns to initialize and update a plan; their work does not use a state-transition model and maps new observations to plan updates. 

\textit{Value prediction networks} (VPNs) by \citet{OhSL17}, \textit{Predictron} by \citet{silver2016predictron}, and \citet{farquhar2017treeqn}, an expansion of VPNs, combine learning and planning by training deep networks to plan through iterative rollouts.
The \textit{Predictron} predicts values by learning an abstract state-transition function. 
VPNs constructs a tree of targets used only for action selection.
\citet{farquhar2017treeqn} create an end-to-end differentiable model that constructs trees to improve value estimates during training and acting. Both \citet{OhSL17} and \cite{farquhar2017treeqn} construct plans using forward-only rollouts by exhaustively expanding each state's actions.
Similarly, \citet{franccois2018combined} proposed a model that combined model-free and model-based components to plan on embedded state representations in a similar fashion to TreeQN \citep{farquhar2017treeqn}. 
They propose an additional loss to the objective function, an approximate entropy maximization penalty, that ensures the expressiveness of the learned embedding.

In contrast to the aforementioned works, during planning DPN learns to selectively expand actions at each state, with the ability to adjust sub-optimal actions, and uses planning results to improve the policy during both training and acting.

\citet{weber2017imagination} proposed Imagination Augmented Agents (I2As), an architecture that learns to plan using a separately trained state-transition model. Planning is accomplished by expanding all available actions $\mathcal{A}$ of the initial state and then performing $\mathcal{A}$ rollouts using a tied-policy for a fixed number of steps. In contrast, our work learns the state-transition model end-to-end, uses a separate policy for planning and acting, and is able to dynamically adjust planning rollouts.
Additionally, in terms of sample efficiency,  I2As require hundreds of millions of steps to converge, with the Sokoban environment taking roughly 800 million steps.
Though not directly comparable, our work in the Push environment, a puzzle game very similar to Sokoban, requires an order of magnitude fewer steps, roughly 20 million, before convergence.

Within continuous control learning a state-transition model for planning has been used in various ways.
\citet{finn2017deep} demonstrate the usage of a predictive model of raw sensory observations with model-predictive control (MPC) where the model is learned in an entirely self-supervised manner. 
\citet{srinivas2018universal} proposed using an embedded differential network that performs iterative planning through gradient descent over actions to reach a specified target goal state within a goal-directed policy. 
\citet{henaff2017model} focus on model-based planning in low-dimensional state spaces and extend their method to perform in both discrete and continuous action spaces.

Additional connections between learning environment models, planning and controls, and other methods related to ours were previously discussed by \citet{schmidhuber2015learning}.

\begin{figure*}[t]
    \begin{subfigure}[b]{0.50\textwidth}
        \begin{subfigure}{\textwidth}
            \includegraphics[width=\textwidth,trim={9.5cm 0 0 0},clip]{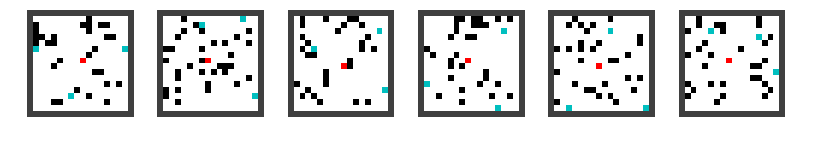}
            \caption{Multi-Goal Gridworld Environment Samples.}
            \label{fig:gw_env_samples}
        \end{subfigure}\\
        \vspace*{\fill}
        \begin{subfigure}{\textwidth}
            \vspace{0.4cm}
            \centering
            \begin{tabular}{c|c}
            \hline
            Model & Avg. Reward \\
            \hline
            DQN-RNN  &  -0.51 \\
            DQN &  -1.26 \\
            A2C &  0.21 \\
            \textbf{DPN-T3} & \textbf{1.3} \\
            \end{tabular}
            \caption{Model Performance.}
            \label{table:model_perf}
        \end{subfigure}
    \end{subfigure}
    \begin{subfigure}[b]{0.5\linewidth}
        \centering
        \raisebox{0.05\height}{
            \includegraphics[width=\textwidth]{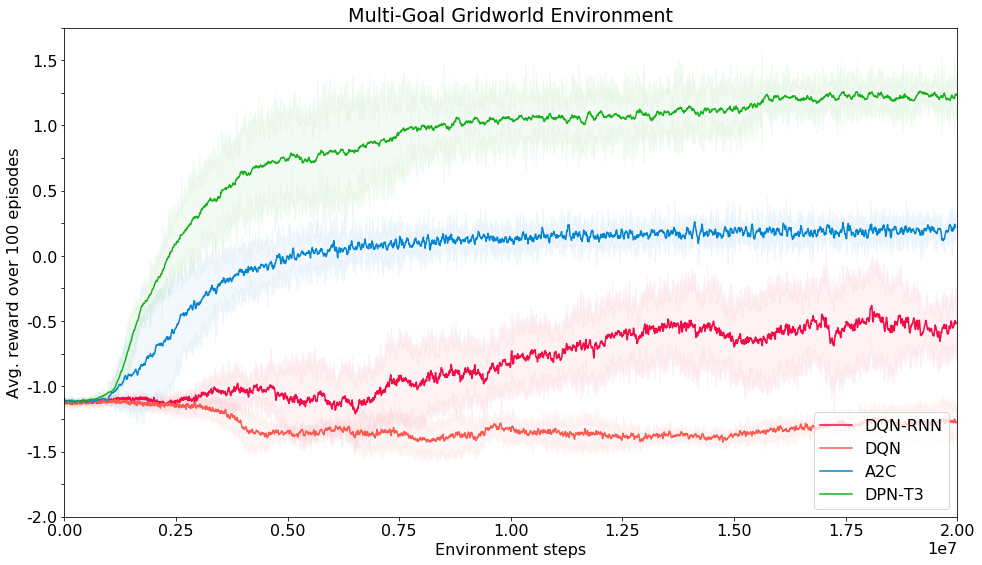}
        }
        \caption{Training Curve.}
        \label{fig:gw_training_curve}
    \end{subfigure}
    \caption{
    \textit{Gridworld Environment.}
    \textit{a)} Randomly generated samples of a $16\times16$ Multi-Goal Gridworld environment where the agent must collect all goals. The agent is shown as red, goals in cyan, obstacles as black, and outside of the environment, not visible to the agent, is shown with a black border. \textit{b)} The performance of each model where \textit{Avg. Reward} is the average of the last 1000 episodes of training. \textit{c)} Training curves with DPN compared to various baselines on $16 \times 16$ Gridworld with 3 goals.
    }
    \label{fig:gridworld_env}
\end{figure*}

\section{Experiments}
\label{sec:experiments}
We evaluated DPN on a multi-goal Gridworld environment and Push, \citep{farquhar2017treeqn} a box-pushing puzzle environment.
Push is similar to Sokoban used by \citet{weber2017imagination} with comparable difficulty.
Within our experiments, we evaluated our model performance against either model-free baselines, DQN and A2C, or planning baselines, such as TreeQN and ATreeC. 
The experiments are designed such that a new scenario is generated across each episode, which ensures that the solution of a single variation cannot be memorized.
We are interested in understanding how well our model can adapt to varied scenarios. 
Additionally, we investigate how planning length $T$ affects model performance, how IA branching during planning affects performance, different distance functions for the IA's reward function, and planning patterns that our agent learned in the Push environment. 
Full details of the environments, experimental setup, hyperparameters are provided in the supplemental material. 
Unless specified otherwise, each model configuration is averaged over 3 different seeds and is trained for 40 million steps.

\textbf{Push}: The Push environment is a box-pushing domain, where an agent must push boxes into goals while avoiding obstacles, with samples shown in Figure \ref{fig:push_environment_samples}.
Since the agent can only push boxes, with no pull actions, poor actions within the environment can lead to irreversible configurations. 
The agent is randomly placed, along with 12 boxes, 5 goals, and 6 obstacles on the center 6x6 tiles of an 8x8 grid. 
Boxes cannot be pushed into each other and obstacles are "soft" such that they do not block movement, but generate a negative reward if the agent or a box moves onto an obstacle. 
Boxes are removed once pushed onto a goal. 
We use the open-source implementation provided by \citet{farquharPushEnv}. 
The episode ends when the agent collects all goals, steps off the map, or goes over 75 steps.
We compare our model performance against planning baselines, TreeQN and ATreeC \citep{farquhar2017treeqn}, as well as model-free baselines, DQN \citep{mnih2015human} and A2C \citep{mnih2016asynchronous}.

\begin{figure*}[t!]
    \centering
    \begin{subfigure}[b]{0.47\textwidth}
        \includegraphics[width=\textwidth]{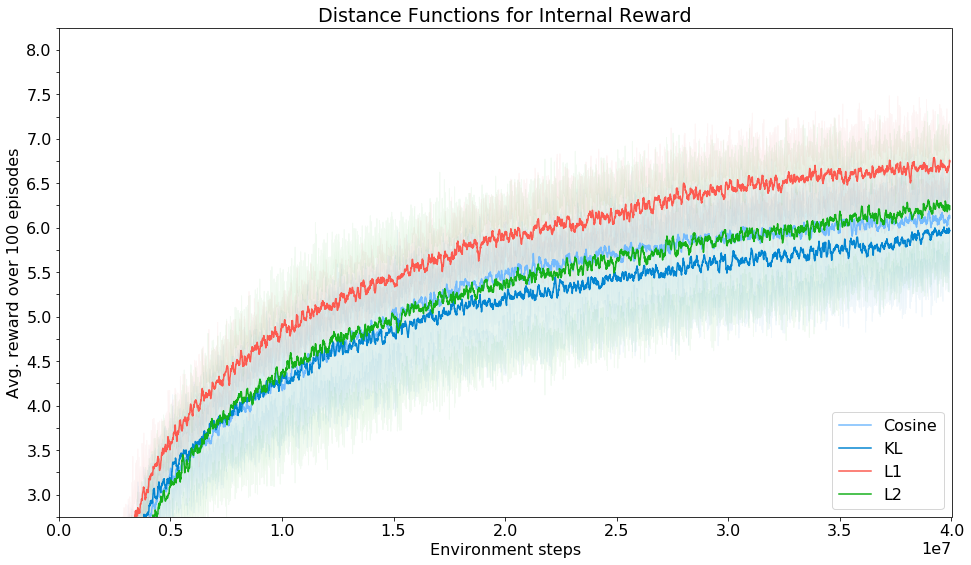}
        \caption{Inner Agent Reward Distance Functions}
        \label{fig:ablation_distance_functions}
    \end{subfigure}
    \hspace*{\fill}
    \begin{subfigure}[b]{0.47\textwidth}
        \includegraphics[width=\textwidth]{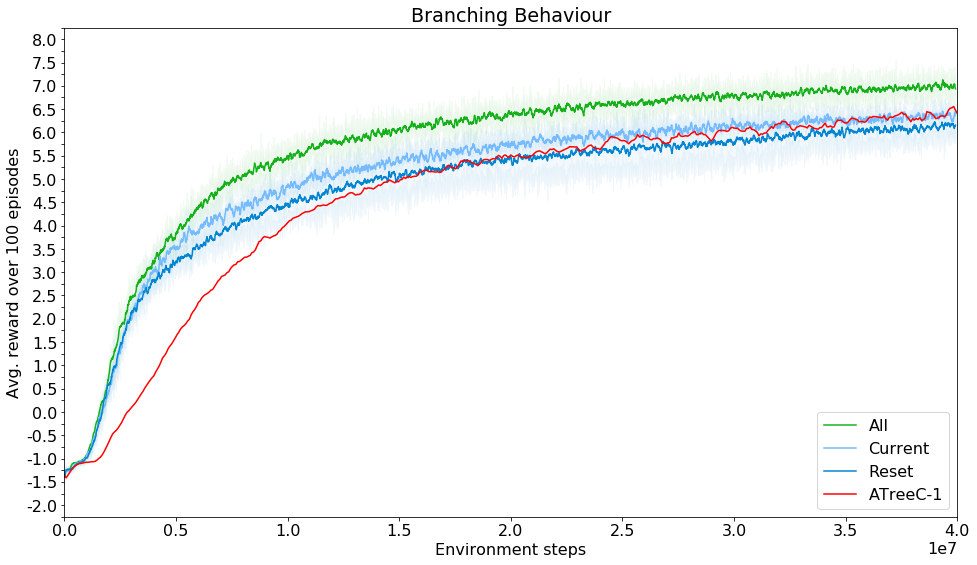}
        \caption{Inner Agent Branching}
        \label{fig:ablation_branching}
    \end{subfigure}
    
    \caption{
     a) \textit{Distance Functions}: the performance of different distance functions. Centered on curve differences.
     b) \textit{Inner Agent Branching}: Various branching choices for IA. \textit{All} corresponds to the default architecture, \textit{Current} results in a forward rollout, and \textit{Reset} is the same as 1-step look ahead. ATreeC-1 corresponds to 1-step look ahead as well.
    }
    \label{fig:ablation_ia}
\end{figure*}

\textbf{Gridworld}: We use a Gridworld domain with randomly placed obstacles that an agent must navigate searching for goals. 
The environment, randomly generated between episodes, is a 16x16 grid with 3 goals.
We force a minimum distance between goals and between the agent and goals.
The agent must learn an optimal policy to solve new unseen maps.
Figure \ref{fig:gw_env_samples} shows several instances of a 16x16 Multi-goal Gridworld. 
The rewards that an agent receives are as follows: +1 for each goal captured, -1 for colliding with a wall, -1 for stepping off the map, -0.01 for each step, and -1 for going over the step limit.
An episode terminates if the agent collides with an obstacle, collects all the goals, steps off the map, or goes over 70 steps. 
We evaluate our algorithm against model-free baselines such as A2C \citep{mnih2016asynchronous} and variants of DQN (recurrent and non-recurrent) \citep{mnih2015human}.
Each baseline used the same encoder structure as DPN.
We train for 20 million environment steps.

\textbf{Planning length}: Using the Push environment, we varied the parameter $T$, which adjusts the number of planning steps, with $T=\{1,2,3\}$ evaluated. 
The Push environment was chosen because the performance is sensitive to an agent's ability to plan effectively.

\textbf{Inner Agent Branching}: We examine the affect on performance of different branching options for $T=3$: current, reset, or all. We also included the ATreeC-1 baseline as this corresponds to the reset branching option of our architecture and serves as a sanity check.

\textbf{Inner Agent Distance Functions}: We vary the distance function used by the IA's loss defined in Equation \ref{eqn:utility}. We examine L1, L2, KL, and Cosine distance functions.

\textbf{Planning Patterns}: We examine the planning patterns that our agent learns in the Push environment for T=3 steps. Here we are interested in understanding what information the agent extracts from the simulation as context before acting.

\section{Results and Discussion}
\label{sec:results}

\subsection{Push Environment}

Figure \ref{fig:push_training_curve} shows DPN, with planning length $T=3$, compared to DQN, A2C, TreeQN and ATreeC baselines \footnote{The data for the training curves of DQN, A2C, TreeQN, and ATreeC were provided by Farquhar \textit{et al.} via email correspondence. Each experiment was run with 12 different seeds for 40million steps.}. 
For TreeQN and ATreeC, we chose tree depths which gave the best performance, corresponding to tree depths of 3 and 1 respectively.
Our model clearly outperforms both planning and non-planning baselines: TreeQN, ATreeC, DQN, and A2C.
We see that our architecture converges at a faster rate than the other baselines, matching ATreeC-1's final performance after roughly 20 million steps in the environment. 
In comparison to the other planning baselines, TreeQN and ATreeC, require roughly 35-40 million steps: $\sim$2x additional samples.

We note that the planning efficiency of DPN is higher in terms of overall performance per number of state-transitions.
On the Push environment, with $\mathcal{A}=4$ actions, TreeQN with tree depth of $d=3$ requires $\big(\frac{\mathcal{A}^{d+1} - 1}{\mathcal{A}-1}\big)-1 = 84$ state-transitions.
In contrast, DPN with planning length of $T=3$ requires only $T$ state-transitions -- a 96\% reduction. Loosely comparing to I2As, simply in terms of state-transitions, we see that I2As require $\mathcal{A} \times L$ state-transitions per action step, where $L$ is the rollout length.  
This performance improvement is a result of DPN learning to selectively expand actions and being able to dynamically adjust previously simulated actions during planning. 

\begin{figure*}[t!]
    \centering
    \begin{subfigure}[h]{0.47\textwidth}
        \includegraphics[width=\textwidth]{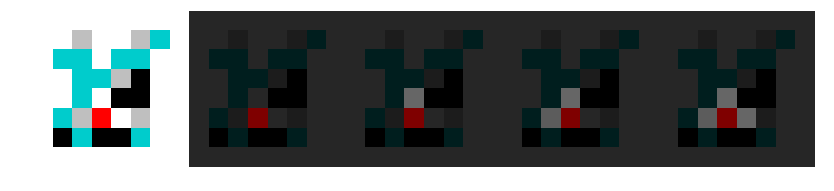}
        \includegraphics[width=\textwidth]{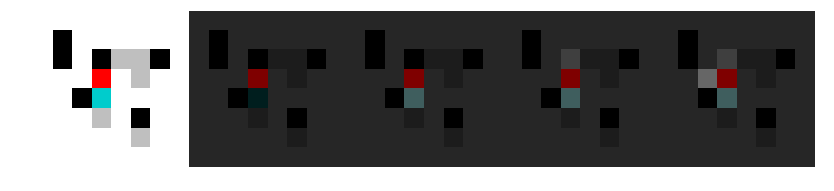}
        \caption{Breadth-first Pattern.}
        \label{fig:push_breadth_planning}
    \end{subfigure}
    \hspace*{\fill}
    \begin{subfigure}[h]{0.47\textwidth}
        \includegraphics[width=\textwidth]{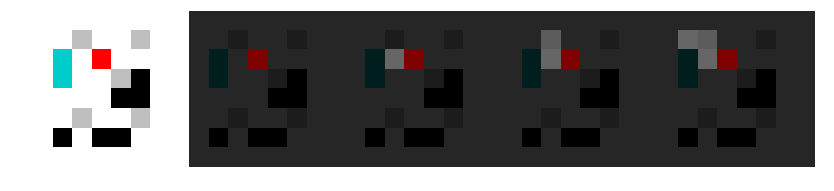}
        \includegraphics[width=\textwidth]{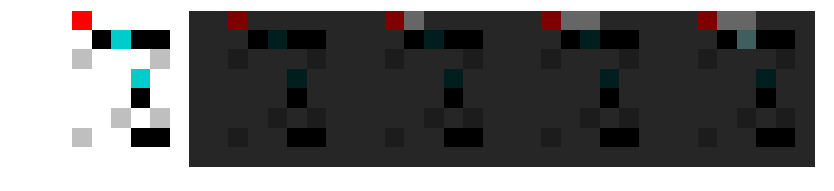}
        \caption{Depth-first Pattern.}
        \label{fig:push_depth_planning}
    \end{subfigure}
    
    \caption{Samples of planning patterns the agent uses to solve the Push environment with $T=3$. The faded environments, to the right of each sample, is used to signify when the agent is planning. Highlighted squares represent the location that IA chose to move towards during planning.}
    \label{fig:push_planning_pattern}
\end{figure*}

\subsection{Multi-Goal Gridworld}

Figure \ref{fig:gw_training_curve} shows, the results of DPN compared to various model-free baselines. 
Within this domain, the difference in performance is clear: our model outperforms the baselines by a significant margin.
The policies that DPN learns generalizes better to new scenarios, can effectively avoid obstacles, and is able to capture multiple goals.
Of the model-free baselines, we see that the A2C baseline performs the best.
We believe that the A2C baseline is able to better explore the environment due to the multiple workers running in parallel throughout training.
Additionally, as seen in Figure \ref{fig:gw_training_curve}, the DQN variants fail to capture any goals and do not achieve a score higher than -1.0. This indicates the DQN baselines learn only to navigate around the map without collecting goals before an episode ends.
It should be noted we saw little performance improvement even when allocating the model-free algorithms an additional 2x environment steps (40 million) or, in the case of DQN models, a 2-4x longer exploration period (8-16 million).

The poor performance of the baseline models might be the result of high variance in the environment's configuration between episodes.
We believe that DPN performs better because it captures common structure present between all permutations of the environment by using the environment model. This allows it to exploit the model for planning in newly generated mazes.

\begin{figure}[H]
    \centering
    \includegraphics[width=0.45\textwidth]{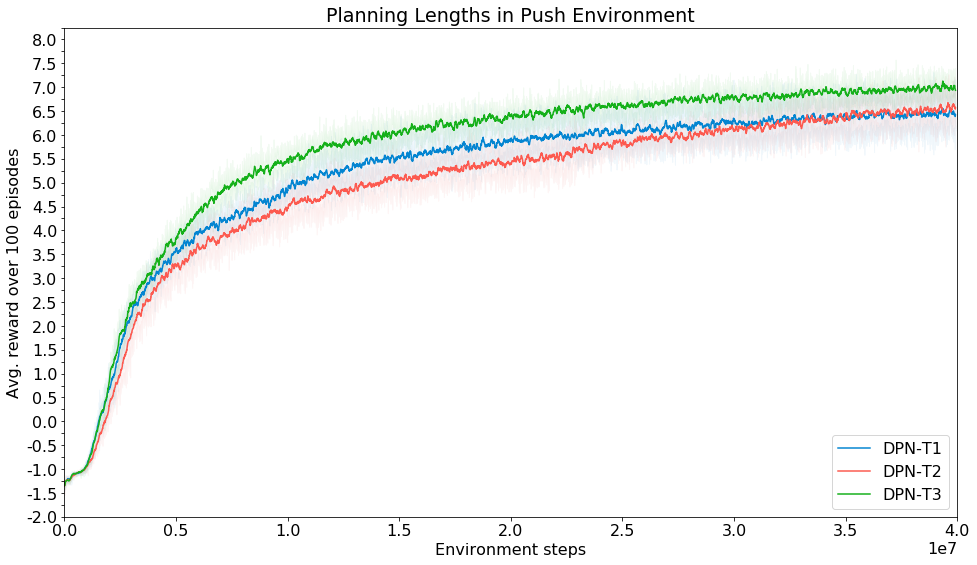}
    \caption{Training over varying planning lengths, $T=\{1,2,3\}$, in the Push Environment. Centered on curve differences.}
    \label{fig:push_planning_curves}
\end{figure}
\subsection{Planning length}
In Figure \ref{fig:push_planning_curves}, we see the performance of our model over the planning lengths $T=\{1,2,3\}$.
As seen in Figure \ref{fig:push_planning_curves}, model performance increases as we add additional planning steps, while the number of model parameters remains constant. 

As the planning length increases, we see the general trend of faster model convergence.
Even a single step $T=1$ of planning allows the agent to test action-hypotheses and avoid poor choices in the environment.
From Figure \ref{fig:push_planning_curves} we see that an additional planning step, from $T=1$ to $T=2$, does not provide benefit until later in training. 
The planning length $T=3$ provides the best performance and faster convergence.
We suspect that the shorter planning lengths do not allow the IA to learn a policy that provides enough utility to OA.
Ideally, the architecture would be able to adjust the number of planning steps $T$ dynamically based on current needs. We see this expansion, similar to the adaptive computation presented by \citet{graves2016adaptive}, as an interesting avenue for future work.

\subsection{Inner Agent Distance Functions}

From \ref{fig:ablation_distance_functions}, we see an evaluation of distance functions used in Equation \ref{eqn:utility}.
The L1 distance function has the best performance with slightly faster convergence.
While the L2, Cosine, and KL functions have worse performance.
We hypothesize that the L1 distance performed better due to its robustness to outliers, a likely event during learning, as the distance is a function of noisy and changing vectors from the OA and state-transition model.

\subsection{Inner Agent Branching}
In Figure \ref{fig:ablation_branching}, we see how different branching options affects the architecture performance. 
Our proposed branching improves performance of the architecture as compared to the \textit{current} and \textit{reset} options.
Interestingly, the performance of our architecture when using the \textit{reset} options is roughly the same as ATreeC-1. This is unsurprising as the \textit{reset} and ATreeC-1 options employ a similar planning strategy of a shallow 1-step look-ahead. 
The small discrepancy in performance could be due to ATreeC-1 evaluating all 4 actions while DPN evaluates only 3.
We see that the \textit{current} branching option results in better performance and is amounts to a forward-only rollout of length $T=3$. We hypothesize the performance difference between \textit{current} and \textit{reset} is from DPN being able to see the results of its actions from the first planning step over a longer time span.

\subsection{Planning Patterns}

By watching a trained agent play through newly generated maps, we identified common planning patterns, which are shown in Figure \ref{fig:push_planning_pattern}.
Two prominent patterns emerged: breadth-first search and depth-first search.
From Figure \ref{fig:push_breadth_planning} we can see that our agent learned to employ breadth-first search, where planning steps are used to expand available actions around the agent, corresponding to a tree of depth 1.
In contrast, depth-first search as seen in Figure \ref{fig:push_depth_planning}, has the agent expanding state forward only. The agent does not always follow depth-first search paths and seems to use them to "check" if a particular pathway is worth pursuing.

\section{Conclusion}

In this paper, we have presented DPN, a new architecture for deep reinforcement learning that uses two agents, IA and OA, working in tandem.
Empirically, we have demonstrated that DPN outperforms the model-free and planning baselines in both the mutli-goal Gridworld and Push environments while using $\sim$2x fewer environment samples.
Ablation studies have shown that our proposed target for the IA improves performance and helps increase the speed of convergence.
We have shown that the IA learns to dynamically construct plans that maximize utility for the OA; with IA learning to dynamically use classical search patterns, such as depth-first search, without explicit instruction. 
Compared to other planning architectures, DPN requires significantly fewer state-transitions during planning for the same level of performance -- drastically reducing computational requirements by up to 96\%.

\section*{Acknowledgments}
We would like to thank Eder Santana, Tony Zhang, and Justin Tomasi for helpful feedback and discussion. This project received funding from Ontario Centres of Excellence, Voucher for Innovation and Productivity (“VIPI”) Program Project \#29393.

\bibliography{icml2019_refs}
\bibliographystyle{icml2019}

\end{document}